# Review Based Entity Ranking using Fuzzy Logic Algorithmic Approach: Analysis

***Pratik N. Kalamkar, Anupama G. Phakatkar***

**Abstract**—Opinion mining, also called sentiment analysis, is the field of study that analyzes people's opinions, sentiments, evaluations, appraisals, attitudes, and emotions towards entities such as products, services, organizations, individuals, issues, events, topics, and their attributes. Holistic lexicon-based approach does not consider the strength of each opinion, i.e., whether the opinion is very strongly negative (or positive), strongly negative (or positive), moderate negative (or positive), very weakly negative (or positive) and weakly negative (or positive). In this paper, we propose approach to rank entities based on orientation and strength of the entity's reviews and user's queries by classifying them in granularity levels (i.e. very weak, weak, moderate, very strong and strong) by combining opinion words (i.e. adverb, adjective, noun and verb) that are related to aspect of interest of certain product. We shall use fuzzy logic algorithmic approach in order to classify opinion words into different category and syntactic dependency resolution to find relations for desired aspect words. Opinion words related to certain aspects of interest are considered to find the entity score for that aspect in the review.

## 1 INTRODUCTION

Opinion mining, also called Sentiment analysis, is the field of study that analyses people's opinions, sentiments, evaluations, appraisals, attitudes, and emotions towards entities such as products, services, organizations, individuals, issues, events, topics, and their attributes [2]. The rapid growth of the social media and the internet has made available lot of opinions regarding certain services, product or any other entity freely available on the internet. Opinions are central to almost all human activities and are key influencers of our behaviors. People’s decisions are motivated by what others think about certain service, product. Current search engines or ranking methods work basically on information retrieval. This is a very active research area. There are several reasons for this. First, it has a wide arrange of applications, almost in every domain. Second, it offers many challenging research problems, which had never been studied before.

While a lot of research recently has been done in the field of opinion mining and some of it dealing with a ranking of entities based on a review or opinion set, classifying opinions into the finer granularity level and then ranking entities based on desired aspect of entities has never been done before. In this paper method for opinion mining from statements at a finer level of granularity using a fuzzy logic approach and then rank entities as per this information is proposed.

Starting with, consider some background of some previous research in field of entity ranking and opinion mining. Then a study of our proposed system in detail is done that tries to make entity ranking a better experience for user by overcoming shortcomings of earlier methods and using some already done good quality research in opinions mining.

## 2 BACKGROUND

Use of opinions in order to rank entities based on analysis of all reviews of the entity and user’s queries at a more granular level is itself challenging [5]. This is so because; Opinion mining faces challenges like irony and sarcasm. Therefore sentiments analysis is a great challenge while analyzing opinions expressed. In such scenario mining opinion at a more detailed level of granularity will be challenging for entity ranking.

In a paper titled “Opinion-Based Entity Ranking” by (Kavita Ganesan & Cheng Xiang Zhai, 2010) [4] given a user's keyword query that expresses the desired features of an entity, and then ranking of all the candidate entities based on how well opinions on these entities match the user's preferences is done. The limitation of this paper being that having not considered the granularity of opinions entity ranking can hardly get more precise. They propose the use of a method called opinion expansion along with the standard BM25 algorithm. In their method there is a lack of resolution of syntactic dependency relating to aspect words. So cannot get more precise ranking.

In another paper titled “Sentiment Classification of Customer Reviews Based on Fuzzy logic" author Samaneh Nadali propose the use of fuzzy logic to classify user opinions into more granular levels [5].

Considering the shortcomings of each of these methods and space of improvement in just before discussed paper, a new system is proposed that combines to give a new system that will be better than normal information retrieval system. How the newly created system behaves under a standard widely used BM25 ranking algorithm will be studied. This is described in sections which follow.

## 3 Proposed System

The proposed system is explained as follows. Ranking of entities is done by first classifying their opinions into finer granular classes and then taking total summarization of its opinion set to match best with user entered query. It contains of three steps which are described as below.

Step: 1) Extraction of aspects using conditional random field machine learning.

Step: 2) Classification of opinions related to aspect word using fuzzy logic algorithmic approach.

Step: 3) Raking of entities so as to best match user preference.

Description of each step in detail is as follows.

### 3.1 Extraction of aspects using Conditional Random Field (CRF) machine learning

First step is to propose methods for aspect extraction. Various methods for aspect extraction are used viz. Extraction based on frequent nouns and noun phrases, Extraction by exploiting opinion and target relations, Extraction using supervised learning, Extraction using topic modeling. Out of this in our proposed system use the one based on supervised learning is used. And within a supervised learning use of the Conditional Random Fields (CRF), hence used as short CRF at some places, a probabilistic model, to find aspects in opinions is made. These are chosen because it will help us to have a learning component, which will make improvements constantly as more data from a variety of domains is trained. On the other hand, if use of hard core methods like those based on frequent nouns, scope of improvement will be limited. Conditional Random Fields for segmenting and labeling sequence data were first proposed by John

Lafferty; these models are probabilistic in nature and can be defined as follows.

Consider, X is a random variable over data sequences to be labeled, and Y is a random variable over corresponding label sequences. All components Yi of Y are assumed to range over a finite label alphabet Y. CRF can be defined as, Let G = (V,E) be a graph such that Y = (Yv)v∈V , so that Y is indexed by the vertices of G. Then (X,Y) is a conditional random field in case, when conditioned on X, the random variables Yv obey the Markov property with respect to the graph: p(Yv |X,Yw,w != v) = p(Yv |X,Yw,w ~ v), where w ~ v means that w and v are neighbors in G [12].

Using above explanation of CRF extraction of the aspect from opinion sentence is done. For doing so a training data set is provided i.e. opinions consisting of desired aspects that are to be extracted from test data and train CRF so that aspects are pointed out from review and also its syntactic dependence is resolved to know what is user's opinion regarding certain aspect. CRF also offers several advantages over HMM.

### 3.2Classification of opinions related to aspect word using fuzzy logic algorithmic approach

After we extract and know the aspect word in sentence next step is to resolve its syntactic dependency to find opinion words related to it and then get scores for these opinion words. Classical logic system works well when there is clear, absolute or mathematical truth. Like is 1+1=2? Answer can take only two values viz. Yes or no. On the other hand, however there are some problems whose answer may depend upon user's perception or whose output may not be clear. Fuzzy logic is a form of many-valued logic; it deals with reasoning that is approximate rather than fixed and exact. Compared to traditional binary sets (where variables may take on true or false values) fuzzy logic variables may have a truth value that ranges in degree between 0 and 1. Fuzzy logic has been extended to handle the concept of partial truth, where the truth value may range between completely true and completely false [13].

Opinions classification problem is mostly same. Opinions are classified as positive, negative or as neutral. But how much positive words in opinions can classify it as positive or how much negative words in opinion can classify it as negative? Use of fuzzy logic in such cases can help us classify opinions into more granular levels of positivity and negativity. This method as follows following steps,

**3.2.1 Finding of opinion words from sentence: -** Finding of opinions words so as to classify opinion's strength. Opinion words are adjective and Adverbs. Use of POS (part of speech tagger) named OpenNLP is done to mark these adjectives and adverbs. Opinion words that feature in describing our aspect of interest are only considered and for this use of Stanford syntactic dependency module is made. For ex. Consider a sentence "The car is good having very stable handling". In this sentence only opinion words "very" and "stable" are considered if user is interested in aspect "handling".

**3.2.2 Fuzzy logic system: -** As discussed earlier fuzzy logic system will be implemented. Here steps will include fuzzification of input where special degree for each of this opinion words is associated use of SentiWords[9]. SentiWords consists of 155000 words having polarities between -1 to 1. However, for our purpose use of only selected 6800 words was made. Then a triangular membership function is designed to divide into three levels low, moderate and high. A fuzzy rule designing is done to find orientation of review. The rules are based on the presence of adverbs, adjectives, verbs and nouns. For ex, If the adverb is high and adjective is high, then orientation is high. Finally, fuzzy results are converted into crisp values using a Memdani's defuzzification function.

**3.2.3 Final output:** - Final orientation along with strength is finally obtained. This is followed for all review sentences that show presence of aspect of interest. This same method for finding orientation and strength of query entered by user will be followed.

### 3.3 Raking of entities so as to best match user preference

The next and most important step of our proposed model is ranking of entities. This is so because hence before research on ranking of entities using opinions that has been done has not considered opinions belonging to certain aspect of an entity along with granular scores that is calculated using fuzzy logic. Our previous steps show that how we have achieved this. For ranking an entity all scores from its all review related to aspect of interest is considered and aggregated, entities are then ranked in descending order of their score.

The proposed algorithm is compared with BM25. BM25 is chosen because it is effective and robust for many tasks.

## 4 Results and Analysis of system

The above proposed method is implemented on a database of cars that consists of about 42,230 reviews belonging to over 150 models of cars from three years. User will enter his desired aspect about a car and ranked list of various cars will be shown with a score based on his entered aspect.

Aspects were extracted by using CRF algorithm, for this purpose 12,000 reviews (about 33 percent of the total reviews available) were manually annotated for aspects in this semi-supervised approach. This is our training data for aspects. Various aspects relating to cars such as mileage, handling, interiors, exteriors, sound system, brakes etc. were trained and extracted using CRF training and testing. After this various steps as mentioned earlier are performed.

Following data shows comparison of proposed system with normal BM25 system. Lemur tool is used to rank dataset using BM25. It can be clearly noticed that the proposed methodology's ranking is totally different from that of BM25 which lacks use of syntactic dependency[7] and fuzzy logic for more granular classification of sentiments.

| Entity Name | Proposed System | | BM25 | |
|---|---|---|---|---|
| | Rank | Score | Rank | Score |
| mazda_rx-8 | 1 | 3.5483 | 8 | -5.818 |
| bmw_6_series | 2 | 2.3656 | 7 | -5.562 |
| suzuki_reno | 3 | 1.8086 | 5 | -5.274 |
| lexus_gs_450h | 4 | 1.3 | 2 | -5.134 |
| chevrolet_malibu_maxx | 5 | 1.1767 | 4 | -5.227 |
| cadillac_escalade_ext | 6 | 1 | 1 | -4.979 |
| chrysler_crossfire | 7 | 0.9451 | 6 | -5.472 |
| volvo_s80 | 8 | 0.848 | 3 | -5.212 |

Table 1: Comparison with standard retrieval system BM25.

Following data shows the accuracy analysis of system. In order to measure how accurately system is calculating in accordance to proposed methodology a standard ideal score for each review file is prepared. This is then compared with score calculated by system.

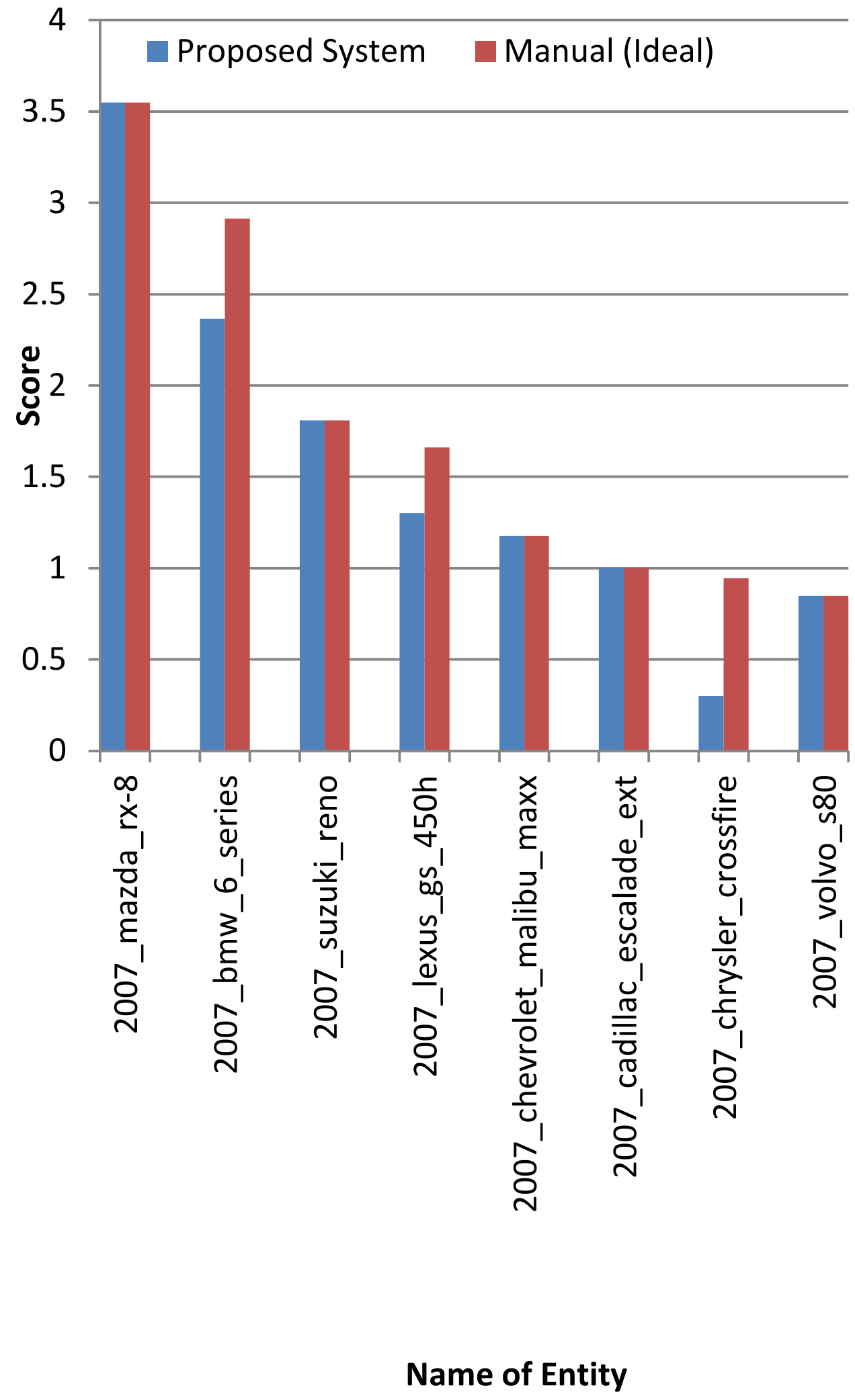

Graph 1: Comparison with ideal score.

The difference between actual expected and calculated score for some entities is due to inability of the system to resolve syntactic dependency, which in turn in some cases is due spelling or grammatically incorrect reviews. In some cases it is due to inability of the system to find scores for opinion words in the dictionary.

Following data shows time and memory usage performance of the proposed system. All measurements were done on Intel multicore processor. Use of multithreading was done intensively through the coding in order to make code more efficient.

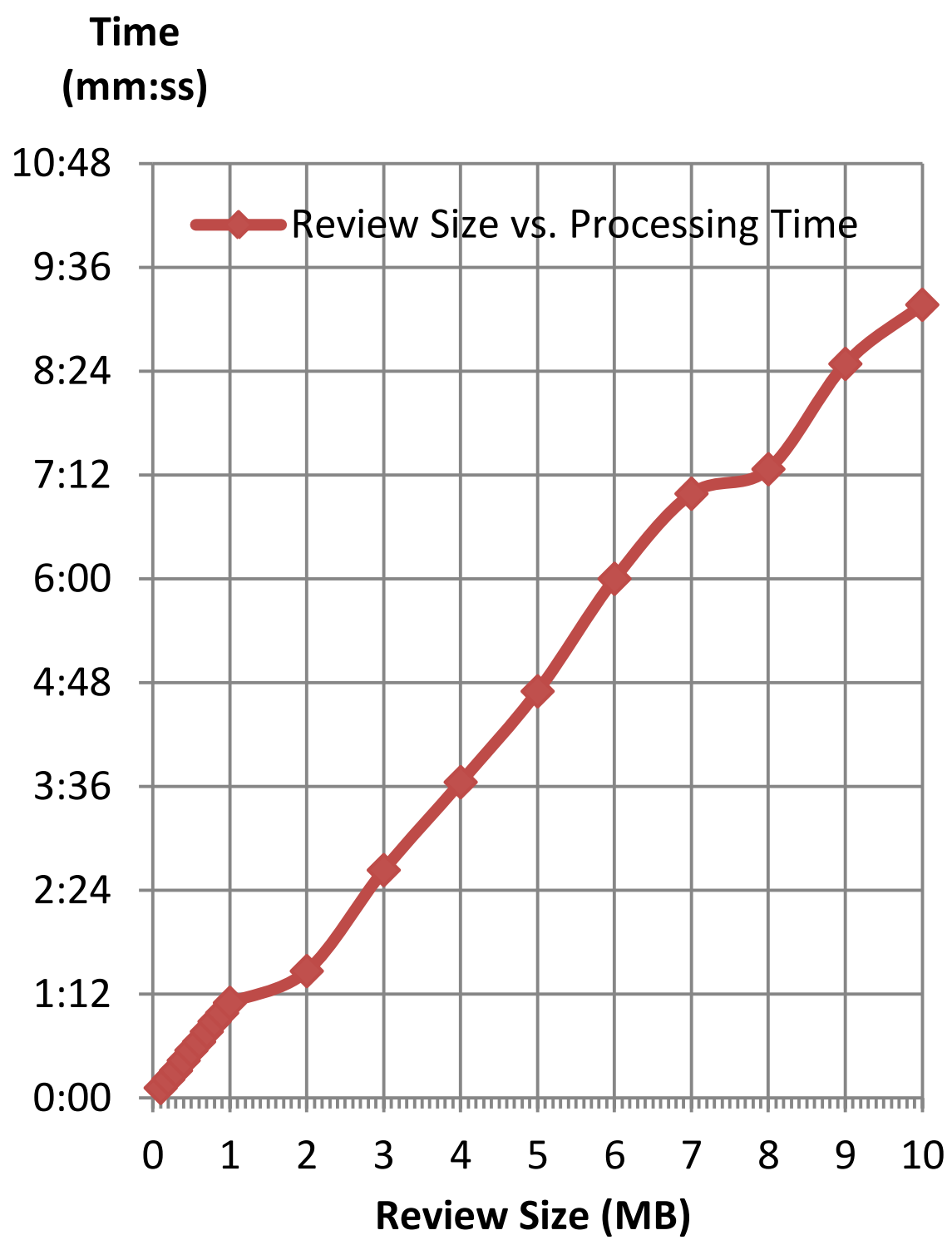

Graph: Review Size vs. Processing Time

We can see processing time increases linearly with increase in size of review dataset.

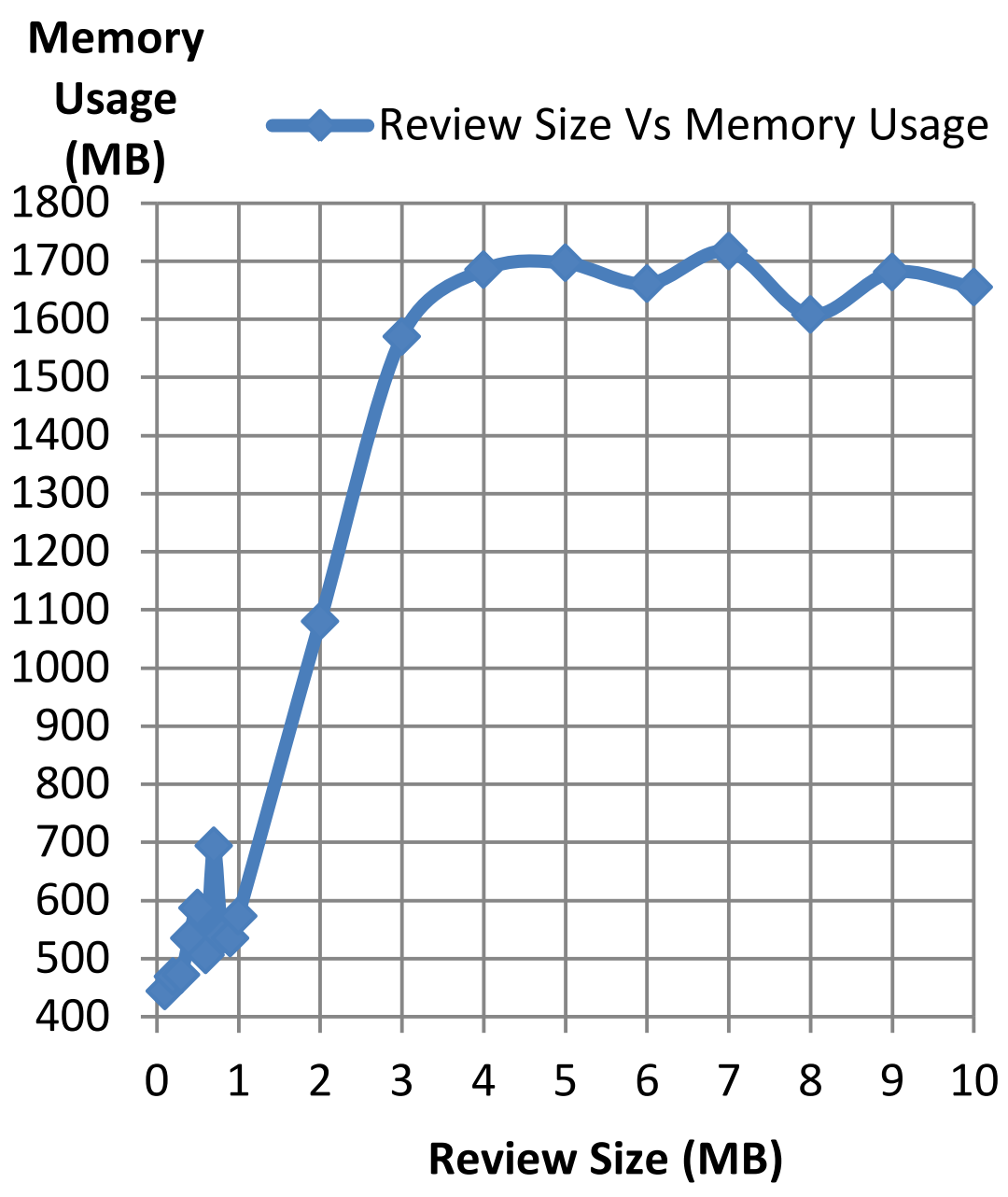

Graph: Review Size vs. Processing Time

Memory usage first increases drastically and then stabilizes, this is due to use of multithreading that tends to use all cores when considerable amount of data is given for processing.

## 5 Conclusion

Use of these methods will greatly enhance the ranking of an entity based on the reviews the entity has and the user query. More precise Ranking than normal information Retrieval is obtained. Use of aspects along with its orientation and strength makes user get precise results. Use of fuzzy logic classifies opinions into more granular level. Ranking based on how well entities aspect satisfiesthe user's query. The proposed system can be extended as an add-on to for a normal search engine like Goggle, Bing, etc. This will help users get more precise and crisp search results, improving usability of a search engine. Also system can be used at online shopping websites to give user better experience of ranked entities as per his entered query.

## Acknowledgment

Authors are thankful to Prof. Girish Potdar, HOD, Computer Department, PICT. for his support and encouragement for given project. We are also thankful to Samaneh Nadadali and Kavita Ganeshan whose previous work provided the foundation for our research project. At last we would like to thank PICT Master of Engineering Department staff for their support.